\documentclass{article}

\usepackage{microtype}
\usepackage{graphicx}
\usepackage{subfigure}
\usepackage{booktabs} 
\usepackage[accepted]{icml2021}

\usepackage{hyperref}

\usepackage{icml2021}
\usepackage{mathrsfs}
\usepackage{amsfonts}
\usepackage{mathtools}
\usepackage{amsthm}
\usepackage{pbox}
\usepackage{wrapfig}
\usepackage{multirow}
\usepackage{psfrag}
\usepackage{color}
\usepackage{soul}
\usepackage{xcolor}
\usepackage{array}

\newcommand{\thickhline}{%
    \noalign {\ifnum 0=`}\fi \hrule height 1pt
    \futurelet \reserved@a \@xhline
}

\graphicspath{{Figures/}}

\DeclareMathOperator*{\argmaxA}{max}

\newcolumntype{?}{!{\vrule width 2pt}}

\icmltitlerunning{SAINT-ACC: Safety-Aware Intelligent Adaptive Cruise Control for Autonomous Vehicles Using Deep Reinforcement Learning}

\begin{document}

\twocolumn[
\icmltitle{SAINT-ACC: Safety-Aware Intelligent Adaptive Cruise Control for Autonomous Vehicles Using Deep Reinforcement Learning}





\begin{icmlauthorlist}
\icmlauthor{Lokesh Das}{to}
\icmlauthor{Myounggyu Won}{to}
\end{icmlauthorlist}

\icmlaffiliation{to}{Department of Computer Science, University of Memphis, TN, United States}

\icmlcorrespondingauthor{Myounggyu Won}{mwon@memphis.edu}

\icmlkeywords{Machine Learning, ICML}

\vskip 0.3in
]



\printAffiliationsAndNotice{}  

\begin{abstract}
We present a novel adaptive cruise control (ACC) system namely SAINT-ACC: \underline{S}afety-\underline{A}ware \underline{Int}elligent \underline{ACC} system (SAINT-ACC) that is designed to achieve simultaneous optimization of traffic efficiency, driving safety, and driving comfort through dynamic adaptation of the inter-vehicle gap based on deep reinforcement learning (RL). A novel dual RL agent-based approach is developed to seek and adapt the optimal balance between traffic efficiency and driving safety/comfort by effectively controlling the driving safety model parameters and inter-vehicle gap based on macroscopic and microscopic traffic information collected from dynamically changing and complex traffic environments. Results obtained through over 12,000 simulation runs with varying traffic scenarios and penetration rates demonstrate that SAINT-ACC significantly enhances traffic flow, driving safety and comfort compared with a state-of-the-art approach.
\end{abstract}

\section{Introduction}
\label{sec:introduction}

An adaptive cruise control (ACC) system enables vehicles to automatically maintain the desired headway distance to the preceding vehicle by using various kinds of sensors such as radar, LiDAR, and cameras~\cite{vahidi2003research}. It is one of the critical components of autonomous vehicles~\cite{wang2020cooperative}. Recent research demonstrates that it can also be used to enhance traffic flow~\cite{ntousakis2015microscopic,delis2016simulation,nikolos2015macroscopic} by adaptively adjusting the inter-vehicle gap in response to dynamically changing traffic conditions~\cite{bekiaris2019feedback,goni2019using,das2020d}.

Latest intelligent ACC systems, however, focus mostly on enhancing traffic flow, overlooking the impact of adaptive adjustment of inter-vehicle gap on driving safety and comfort~\cite{bekiaris2019feedback,goni2019using}. As such, developing an intelligent ACC system that optimizes not only traffic flow but also driving safety and comfort is an important research problem. Recently, a surrogate safety measure (SSM)~\cite{nadimi2020calculating} such as the time-to-collision (TTC) model~\cite{gettman2003surrogate} is incorporated in the design of an intelligent ACC system to account for driving safety and comfort in making a control decision on the inter-vehicle gap~\cite{zhu2020safe,alizadeh2019automated}. However, we demonstrate that state-of-the-art safety-aware ACC systems~\cite{zhu2020safe,alizadeh2019automated} do not perform driving safety assessment effectively under dynamically changing and complex traffic environments~\cite{zhu2020safe} because the implications of dynamic adaptation of the critical parameters of the driving safety model on balancing traffic efficiency and safety are not fully considered.

In this paper, we develop the \underline{S}afety-\underline{A}ware \underline{Int}elligent \underline{ACC} system (SAINT-ACC) that performs driving safety assessment more effectively through dynamically updating the safety model parameters in response to varying traffic conditions. A novel dual reinforcement learning (RL) agent approach is designed aiming to maximize traffic efficiency, driving safety, and driving comfort simultaneously. Specifically, a separate RL agent is designed to find and adapt the optimal TTC threshold~\cite{gettman2003surrogate} based on rich traffic information including both macroscopic and microscopic traffic data obtained from the surrounding environment. This RL agent, by providing the optimal TTC threshold as feedback, interacts with the main RL agent designed to derive an optimal inter-vehicle gap that maximizes traffic flow, driving safety, and comfort at the same time.

The dual RL agents are trained under various highway scenarios including highways with on-ramps and off-ramps. A simulation platform based on the combination of a traffic simulator, SUMO~\cite{krajzewicz2012recent}, and a vehicular network simulator, Veins~\cite{sommer2019veins}, is used to perform extensive simulations over 12,000 runs with varying penetration rates and merging/exiting traffic density to evaluate the performance. Results demonstrate the superior performance of the first `dynamic' driving safety model based on RL that assesses driving safety more effectively under complex traffic conditions compared to the existing ‘static’ TTC model of the state-of-the-art intelligent ACC systems~\cite{zhu2020safe} and the applicability of SAINT-ACC for various transportation applications such as truck platooning and traffic control.

\section{Related Work}
\label{sec:related_work}

The methods for assessing driving safety for autonomous vehicles are categorized largely into two groups: rule-based and machine learning-based approaches. The rule-based methods are based on a safety model represented as a function to gauge the level of driving safety. These methods are easy to implement and are capable of assessing driving safety very quickly, thereby being adequate for fast-moving vehicles that should make a control decision quickly. Nilsson \emph{et. al.} design a utility function to determine whether or not it is safe for an autonomous vehicle to change lanes~\cite{nilsson2016if}. The utility function computes a utility value based on the average vehicle speed, traffic density on a lane, and the remaining time until the vehicle reaches the end of the lane; If the utility value of a target lane is greater than that of the current lane, changing lanes is considered safe. Zheng \emph{et al.} develop a safety model based on a threshold for the deceleration of a lane-changing vehicle and the immediately following vehicle~\cite{zheng2019cooperative}. If the deceleration of a vehicle is greater than the threshold, it is considered unsafe to change lanes. Numerous works utilize the surrogate safety measures (SSM) to evaluate the safety impacts of autonomous vehicle technologies~\cite{morando2018studying,virdi2019safety,rahman2018longitudinal}. However, these rule-based approaches do not cope well with dynamically changing traffic conditions~\cite{ye2020automated}.

To address the limitation of the rule-based methods, machine learning techniques are utilized for assessing driving safety~\cite{zhu2020safe}. Numerous research has demonstrated that machine learning-based approaches outperform rule-based methods in evaluating driving safety~\cite{ye2020automated}. Mirchevska \emph{et al.} adopt RL integrated with a safety verification model to assure that the vehicle action is guaranteed to be safe at any time~\cite{mirchevska2018high}. However, their method is designed to make only a binary decision, \emph{i.e.,} whether taking an action is safe or not. Hoel \emph{et al.} adopt a deep Q-Network to allow vehicles to make a lane-changing decision safely~\cite{hoel2018automated}. However, a simple reward function is used considering only extreme cases, \emph{i.e.,} providing a penalizing reward only when a crash occurs; As a result, it does not evaluate driving safety effectively. Similar RL-based approaches are proposed, \emph{e.g.,} Lin \emph{et al.} utilize deep RL for driving safety in training a merging vehicle to minimize collisions~\cite{lin2019anti}; Baheri \emph{et al.} incorporate a rule-based model with a dynamically-learned safety module based on RL~\cite{baheri2019deep}. However, the reward function of these RL-based solutions takes into account only particular dangerous scenarios in giving penalizing rewards, \emph{e.g.,} when merging vehicles stop suddenly or collide. To address this problem, Ye \emph{et al.} develop a solution based on proximal policy optimization (PPO) using deep RL~\cite{ye2020automated}. Specifically, the reward function incorporates the `near-collision' penalty instead of relying only on the situations where a collision actually occurs. Zhu \emph{et al.} design a better reward function based on human driving data and various other performance criteria such as safety, efficiency, and driving comfort~\cite{zhu2020safe}. In particular, to evaluate driving safety, a surrogate safety measure called the time to collision (TTC) is incorporated with the reward function. However, a fixed threshold for TTC is used in their model (\emph{i.e.,} 4 seconds), constraining the performance under dynamically changing traffic conditions. Similarly, Alizadeh \emph{et al.} develop a deep RL agent based on TTC to enable autonomous vehicles to change lanes safely~\cite{alizadeh2019automated}. However, it also relies on a fixed TTC threshold (1.8 seconds). In contrast to these state-of-the-art research works, we develop a novel dual RL agent approach where the optimal TTC threshold is determined adaptively depending on dynamically changing traffic conditions, which is provided as feedback to the main RL agent for controlling the inter-vehicle gap more effectively.

\section{Motivation}
\label{sec:motivation}

A surrogate safety measure (SSM) is a widely used indicator of driving safety~\cite{nadimi2020calculating}. There are different kinds of SSMs such as time to collision (TTC), unsafe density (UD), proportion of stopping distance (PSD), comprehensive time-based measure (CTM), \emph{etc.}~\cite{guido2011comparing}. The most widely used one is TTC that represents the amount of time remaining until a collision with the front vehicle will occur if the current vehicle keeps its trajectory and speed~\cite{gettman2003surrogate}. More specifically, TTC is based on a threshold denoted by TTC* defined to determine whether or not a collision will occur; Formally, the TTC of vehicle $F$ at time $t$ denoted by $\mbox{TTC}_F(t)$ with respect to the preceding vehicle $L$ is defined as follows: $\mbox{TTC}_F(t) = \frac{X_L(t) - X_F(t) - l_L}{v_F(t) - v_L(t)}, \forall v_F(t) > v_L(t)$, where $X$, $v$, and $l$ are the position, speed, and body length of a vehicle, respectively. The TTC model is only valid when the speed of vehicle $F$ is greater than that of the leading vehicle $L$ (\emph{i.e.,} $v_F(t) > v_L(t)$). A dangerous situation is detected at time $t$ if the measured TTC of a following vehicle is smaller than the threshold $\mbox{TTC}^*$ at time $t$ (\emph{i.e.,} $\mbox{TTC}_F(t) < \mbox{TTC}^*$), which indicates that the effectiveness of the TTC model depends on how the threshold $\mbox{TTC}^*$ is determined.

An interesting observation is that state-of-the-art intelligent ACC systems designed based on the TTC model for evaluating driving safety rely on a fixed TTC threshold (TTC$^*$)~\cite{zhu2020safe,alizadeh2019automated}. In this section, we perform a simulation study to understand the effect of TTC$^*$ on the performance of an intelligent ACC system aiming to motivate the needs for a novel method to adapt TTC$^*$ dynamically in response to the current traffic conditions.

We implement a state-of-the-art intelligent ACC system based on RL integrated with the TTC model~\cite{zhu2020safe}. More specifically, in contrast to~\cite{zhu2020safe} built upon a fixed TTC$^*$ (\emph{i.e.,} 4 seconds), in this study, we attempt to vary TTC$^*$ to demonstrate the effect of the threshold on the performance of the ACC system in terms of driving safety and traffic efficiency. A highway segment with an on-ramp is created for this simulation study. The details of the simulation settings are described in Section~\ref{sec:simulation_results}. The driving safety is represented as the number of `near' collisions. Here a near collision is defined as a situation where there is a vehicle with the inter-vehicle gap smaller than a minimum gap (default 2.5m)~\cite{krajzewicz2012recent}. Additionally, it is ensured that if the speed of a vehicle is smaller than or equal to that of the leading vehicle, a near collision is not registered even if the inter-vehicle gap is smaller than the minimum gap. The traffic efficiency is represented by the average vehicle speed of all vehicles on the given highway segment.

\begin{figure}[h]
\centering
\begin{minipage}{.49\columnwidth}
  \centering
  \includegraphics[width=\linewidth]{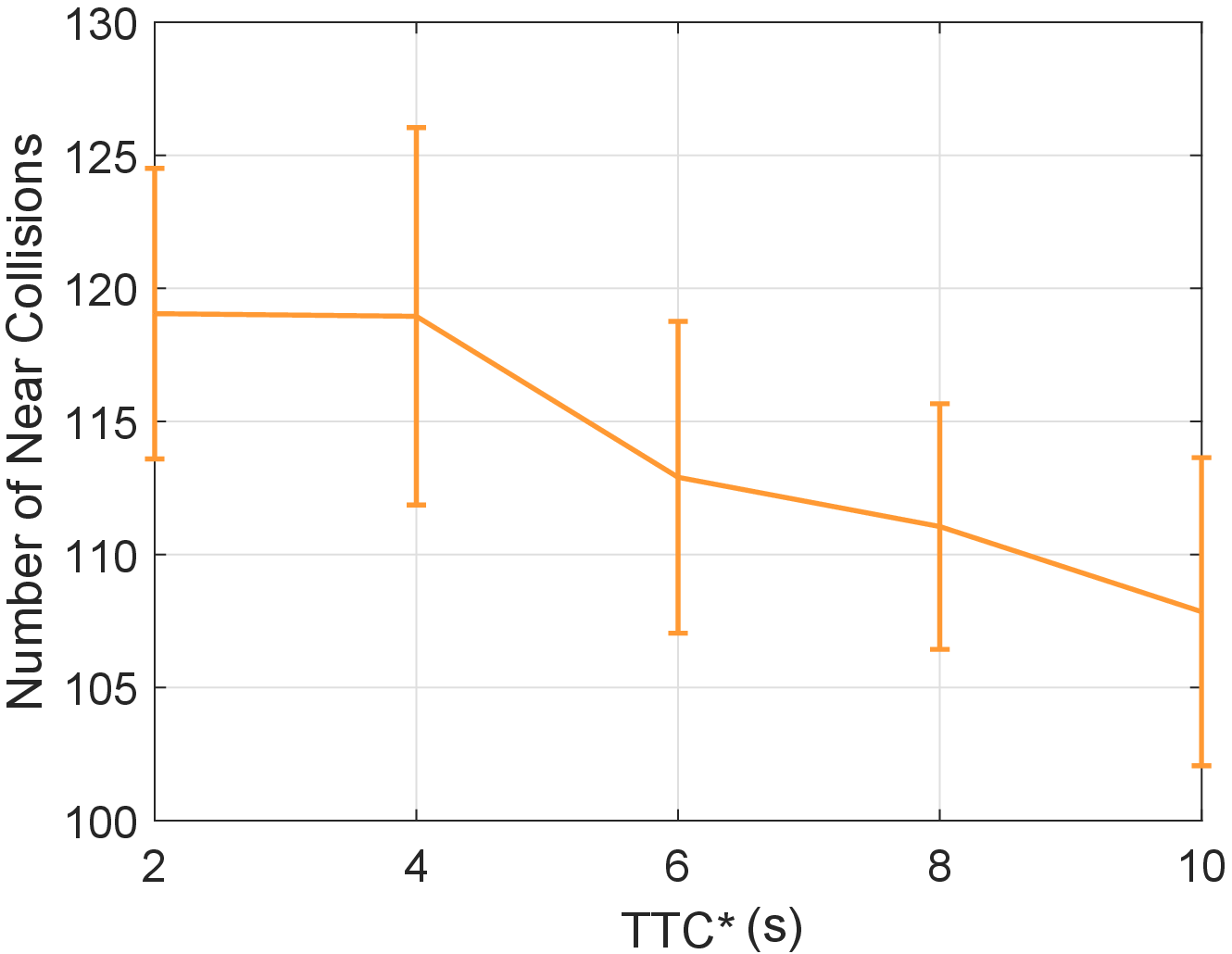}
  \vspace{-15pt}
  \caption{The effect of TTC$^*$ on driving safety.}
  \label{fig:motivation}
\end{minipage}%
\hspace*{3mm}
\begin{minipage}{.49\columnwidth}
  \centering
  \includegraphics[width=\linewidth]{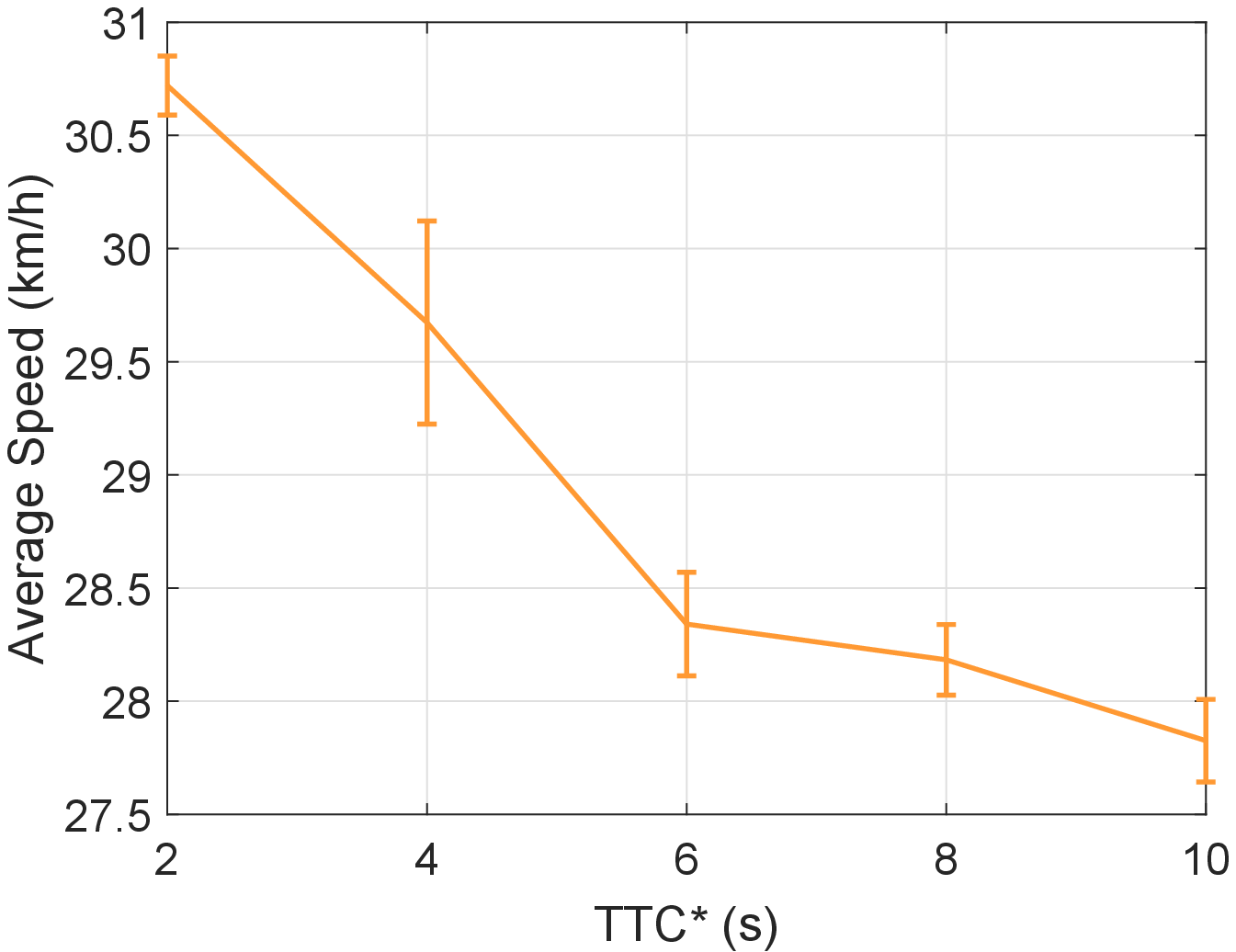}
  \vspace{-15pt}
  \caption{The effect of TTC$^*$ on traffic efficiency.}
  \label{fig:motivation_speed}
\end{minipage}
\end{figure}

Fig.~\ref{fig:motivation} exhibits the measured driving safety as a function of TTC$^*$. We observe that there exists a tradeoff between driving safety and traffic efficiency depending on TTC$^*$. More specifically, as TTC$^*$ decreases, driving safety degrades; the reason is that some dangerous situations are not detected because TTC$^*$ is too small - recall that a dangerous situation is detected when the TTC value for a vehicle is smaller than TTC$^*$; in contrast, as depicted in Fig.~\ref{fig:motivation_speed}, traffic efficiency increases as TTC$^*$ becomes smaller. Another interesting observation is that when TTC$^*$ is too small, the performance gain in terms of driving safety becomes marginal. The results indicate that there exists a TTC$^*$ that strikes the optimal balance between driving safety and traffic efficiency and that traffic efficiency can be improved further while sustaining a similar level of driving safety. The results motivate us to develop an adaptive intelligent ACC system that adapts TTC$^*$ dynamically to achieve the optimal balance between traffic efficiency and driving safety.

\section{Safety-Aware Intelligent ACC System (SAINT-ACC)}
\label{sec:proposed_approach}



\begin{figure}[h]
\centering
\includegraphics[width=.95\columnwidth]{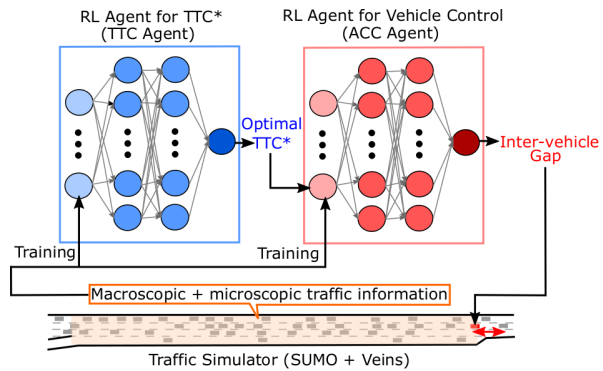}
\caption {An overview of SAINT-ACC.}
\label{fig:overview}
\end{figure}

An overview of SAINT-ACC is depicted in Fig.~\ref{fig:overview}. It is built upon a feedback loop comprising of dual RL agents that is used to adapt TTC$^*$ in response to dynamically changing traffic conditions and in turn utilizes the obtained optimal TTC$^*$ to control the inter-vehicle gap to maximize performance. The TTC agent is designed to obtain the optimal value of TTC$^*$, and the ACC agent is used to control the inter-vehicle gap. Both RL agents are trained based on the macroscopic traffic information including the vehicle density and average vehicle speed as well as the microscopic traffic information such as the vehicle length and vehicle acceleration collected from the surrounding traffic environment. More specifically, the TTC agent interacts with the ACC agent continually by providing the optimal value of TTC$^*$ so that the ACC agent can control the inter-vehicle gap based on the latest and accurate driving safety assessment to improve traffic efficiency and driving safety simultaneously. The dual RL design of SAINT-ACC allows for more effective exploration and improved learning with a smaller action set~\cite{zahavy2018learn}, \emph{i.e.,} we observe that our state space increases by 25X when it is designed with a single consolidated RL agent, and the design of the reward function becomes more complicated with reduced efficiency as it should evaluate heterogeneous performance criteria.

\subsection{Designing the TTC Agent}
\label{subsec:paramters}

The TTC agent is used to adjust TTC$^*$ dynamically in response to the current traffic conditions. This section explains the details of the RL model for the TTC agent.

\begin{wrapfigure}{r}{0.5\columnwidth}
\vspace{0pt}
  \begin{center}
    \includegraphics[width=\linewidth]{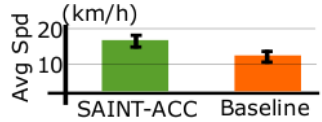}
  \caption{The state space of SAINT-ACC modeling the traffic of a roundabout.}
  \label{fig:roundabout}
  \end{center}
  \vspace{-10pt}
\end{wrapfigure}

\textbf{State Space:} The state space is designed to represent the current traffic conditions. It incorporates both the macroscopic and microscopic traffic information including most of the widely used traffic parameters in the literature, \emph{i.e.,} vehicle acceleration/deceleration, time headway, driving imperfection (according to SUMO~\cite{krajzewicz2012recent}), minimum inter-vehicle gap, vehicle length, mainline vehicle density, mainline average vehicle speed, non-mainline vehicle density, non-mainline average vehicle speed, and the length of ramp (if applicable). A salient aspect of the state space model is that it is designed to consider the traffic parameters of the mainline and non-mainline traffic separately so that SAINT-ACC can be applied to more diverse highway types. For example, while SAINT-ACC is primarily designed for highways with on/off ramps, we demonstrate that it can be applied to other road types such as a roundabout. We create a roundabout scenario and apply SAINT-ACC by modeling the traffic in terms of the mainline traffic and merging traffic. SAINT-ACC allows vehicles in the roundabout to create a gap to enable merging vehicles to enter the roundabout more smoothly without reducing the speed significantly, thereby improving the average vehicles speed compared with a baseline algorithm (Fig.~\ref{fig:roundabout}).

\begin{wrapfigure}{r}{0.5\columnwidth}
\vspace{-18pt}
  \begin{center}
    \includegraphics[width=\linewidth]{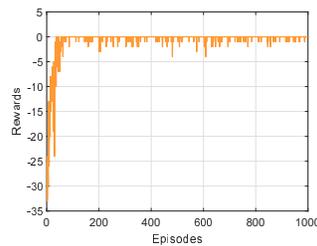}
    \vspace{-15pt}
  \caption{Convergence of the reward function of the TTC agent.}
  \label{fig:reward_rl1}
  \end{center}
  \vspace{-10pt}
\end{wrapfigure}

\textbf{Action Space:} Since the goal of the TTC agent is to find the optimal TTC$^*$ and adapt it dynamically depending on real-time traffic conditions, the action space is primarily designed to control TTC$^*$. A total of 21 actions are defined with respect to a wide range of values for TTC$^*$ between 0 and 10 with an interval of 0.5. We note that TTC$^*$ greater than 10 is not useful as 10 seconds is enough time to avoid a collision in most highway scenarios~\cite{vogel2003comparison}.

\textbf{Reward Function:} The reward function denoted by $R_{TTC}$ is a core part of the TTC agent. It is used to evaluate the effectiveness of the selected TTC$^*$. The reward function is defined based on the concept of a `near' collision (as explained in Section~\ref{sec:motivation}) and actual collisions in simulation. More specifically, the reward function determines the reward values based on the number of false positives (FP), number of false negatives (FN), and number of actual collisions (AC). A false positive is recorded in the event that the TTC model with a given TTC$^*$ detects a dangerous situation (\emph{i.e.,} TTC $<$ TTC$^*$) and a near-collision is not registered in simulation. On the other hand, a false negative is recorded when the TTC model with a given TTC$^*$ does not detect a dangerous situation (\emph{i.e.,} TTC $\ge$ TTC$^*$) and a near-collision is registered. An actual collision is recorded in the event that two cars actually crash in simulation. The reward function $R_{TTC}$ is then calculated as follows: $R_{TTC} = - \alpha_1 (\mbox{Total Number of FPs}) - \alpha_2(\mbox{Total Number of FNs}) - \alpha_3(\mbox{Total Number of ACs})$. We determine those parameters as $\alpha_1 = 1$, $\alpha_2 = 2$, and $\alpha_3 = 10$ through numerous simulations. Note that a highest priority is placed on the number of actual collisions ($\alpha_3$).


\begin{table}[h]
\caption{Parameters used for the neural network of the TTC Agent.}
\vspace{-15pt}
\label{tab:param_network}
\vskip 0.15in
\begin{center}
\begin{tabular}{p{3cm}|p{4.5cm}}
\toprule
Neural network architecture & Input size 13 (state size), output size 21 (action size), 1 hidden layer with 30 neurons in each layer, ``random\_normal'' for kernel initialization\\
\hline
Activation functions & ReLU for hidden layer, ``linear'' for output layer \\
\hline
Replay buffer size & 100k samples \\
\hline
($\gamma, \epsilon_0, \epsilon_{min}$, $\lambda_{decay})$ & (0.95, 1.0,0.01,0.99985) \\
\hline
Batch size & 64  \\
\hline
Loss function & Mean square error (MSE) \\
\hline
Optimization method & Adam with learning rate 0.0001  \\
\hline
Target network update frequency & 5 episodes (each episode runs for 7 minute of simulation time)  \\
\hline
Input Normalization & Batch normalization  \\
\bottomrule
\end{tabular}
\end{center}
\vskip -0.1in
\end{table}

\textbf{Neural Network:} The neural network used to train the TTC agent consists of one hidden layer with 30 neurons. The linear activation function is used for the output layer to allow the deep Q network to produce the Q value with no specific bounds, not necessarily meaning no activation function as the weights are involved. The parameters for the neural network (summarized in Table~\ref{tab:param_network}) are carefully selected based on the extensive trial and error process. Especially, we adopt the $\epsilon$-greedy policy~\cite{wunder2010classes} to balance between exploration and exploitation in training our neural network. An action is randomly chosen from the action space with probability $\epsilon$, while an action is selected using the greedy method with probability $1 - \epsilon$. It is ensured that $\epsilon$ is decreased gradually as the greedy algorithm iterates, \emph{i.e.,} $\epsilon = \argmaxA(\epsilon_0 \cdot \lambda_{decay}, \epsilon_{min})$. The values selected for these parameters $\epsilon_0$, $\lambda_{decay}$, $\epsilon_{min}$ for training our neural network are summarized in Table~\ref{tab:param_network}. Fig.~\ref{fig:reward_rl1} displays the rewards for the TTC agent measured over a large number of episodes, demonstrating the convergence of the reward function.

\subsection{Designing the ACC Agent}
\label{subsec:acc_system}

The ACC agent is trained to control the optimal inter-vehicle gap to maximize traffic flow, driving safety, and driving comfort. The TTC$^*$ value provided by the TTC agent is used in the reward function of the ACC agent to make the control decision.

\textbf{State Space:} The state space of the ACC agent is designed similarly as that of the TCC agent, \emph{i.e.,} the state space represents the current traffic conditions based on macroscopic and microscopic traffic information collected from the surrounding traffic environment.

\textbf{Action Space:} The action space of the ACC agent is designed to represent a set of control actions performed by a vehicle to adjust the inter-vehicle gap. Specifically, the control action is to set the inter-vehicle gap to a certain value selected from a predefined set of available values for the inter-vehicle gap. A total of 25 different actions corresponding to the headway gaps between 1m and 25m with an interval of 1m are used. A vehicle takes an action of adjusting the inter-vehicle gap every second in accordance with the current traffic conditions. This dynamic adaptation of the inter-vehicle gap is particularly useful for highways with merging/exiting traffic because creating an appropriate inter-vehicle gap is crucial for merging and exiting vehicles to change lanes smoothly without causing traffic perturbation that can lead to formation and propagation of a traffic jam. Such a positive effect of adjusting the inter-vehicle gap on preventing a traffic jam has been well studied in the literature~\cite{goni2019using}.


\textbf{Reward Function:} The reward function of the ACC agent denoted by $R_{ACC}$ consists of three subfunctions representing driving safety $R_s$, traffic efficiency $R_e$, and driving comfort $R_c$, respectively. In designing the subfunction for driving safety, we employ a TTC-based approach of the state-of-the-art intelligent ACC system~\cite{zhu2020safe} which measures driving safety based on the TTC model as follows: $F_{TTC} = \begin{cases} log(\frac{TTC}{4}) & \mbox{if } 0 \le TTC \le 4 \\ 0 & \mbox{otherwise} \end{cases}$. An interesting aspect is that the TTC threshold TTC$^*$ is fixed to 4 in this model. In contrast, we make two critical modifications to integrate with the TTC agent: (1) we incorporate the optimal TTC$^*$ which is dynamically adjusted by the TTC agent instead of relying on a fixed value to perform assessment of driving safety more effectively, and (2) we consider not only the driving safety for the ego vehicle but also the driving safety for surrounding vehicles in a given highway segment as a whole. Consequently, our subfunction for driving safety is calculated as follows: $R_{s} = \begin{cases} \sum_{i \in V} log(\frac{TTC_i}{TTC^*}) & \mbox{if } 0 \le TTC \le TTC^* \\ 0 & \mbox{otherwise} \end{cases}$, where $TTC_i$ is the TTC of an individual vehicle $i$, and $V$ is a set of vehicles in the given highway segment.

The subfunction for traffic efficiency ($R_e$) is calculated based on the degree of traffic flow which is measured using the average delay to pass a given highway segment. More precisely, if the average delay is greater than the expected delay for congested traffic conditions (which is known priori), a penalizing reward is imposed. On the other hand, if the average delay is smaller than the delay for congested traffic conditions, a positive reward is provided. Consequently, the subfunction for traffic efficiency is defined as follows: $R_e =
\begin{cases}
      +1 & \mbox{if avg delay} \leq \frac{\mbox{highway length}}{\mbox{avg speed under congestion}} \\
      -1 & \mbox{if avg delay} > \frac{\mbox{highway length}}{\mbox{avg speed under congestion}}
   \end{cases}$.

Lastly, the subfunction for driving comfort ($R_c$) is defined based on an existing driving comfort model~\cite{zhu2020safe,jacobson1980models}. In this driving comfort model, the level of driving comfort is quantified using the change rate of acceleration called Jerk as follows: $R_{c} = - \frac{\mbox{jerk}^2}{27.04}$. Here, since our sampling interval is one second (\emph{i.e.,} vehicles make a control decision on the inter-vehicle gap every second), and the acceleration is bounded between -2.6 and 2.6 m/s$^2$, the greatest possible jerk is $\frac{2.6-(-2.6)}{1.0} = 5.2$. As such, to bound the result of $R_{c}$ between 0 and 1, we divide by 27.04 which is the square of the greatest possible jerk.

Now combining all three subfunctions, the reward function $R_{ACC}$ is defined as follows: $R_{ACC} = \beta_{1}R_e + \beta_{2}R_s + \beta_{3}R_c$, where, $\beta_1, \beta_2$, and $\beta_3$ are weighting parameters which can be used to allow the user to vary the weight of a certain performance criteria. In our simulation, we use $\beta_1 = \beta_2 = \beta_3 = 1$ to optimize traffic efficiency, driving safety, and driving comfort with the same weight.

\begin{wrapfigure}{r}{0.5\columnwidth}
\vspace{-10pt}
  \begin{center}
    \includegraphics[width=\linewidth]{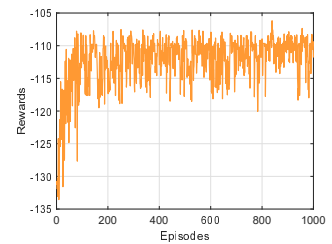}
  \vspace{-15pt}
  \caption{Convergence of the reward function of the ACC agent.}
  \label{fig:reward_rl2}
  \end{center}
  \vspace{-10pt}
\end{wrapfigure}

\textbf{Neural Network:} The neural network of the ACC agent is designed similarly as that of the TTC agent. However, through extensive trial and error, we observe that the most effective neural network configuration for the ACC agent is slightly different as follows: We use one hidden layer with 30 neurons; the replay buffer size is 100k samples; and the batch size is 64. We also adopt the $\epsilon$-greedy policy~\cite{wunder2010classes} for training the neural network for the ACC agent with a slightly smaller $\lambda_{delay}$. Fig.~\ref{fig:reward_rl2} shows the reward values as a function of episodes demonstrating the convergence of rewards for the ACC agent.

\section{Results}
\label{sec:simulation_results}

We implement SAINT-ACC and a state-of-the-art intelligent ACC system (referred to as `Zhu' hereafter)~\cite{zhu2020safe} on a traffic simulation framework, SUMO~\cite{krajzewicz2012recent}, which is integrated with a vehicular network simulator, Veins~\cite{sommer2019veins}. The state-of-the-art intelligent ACC system is based on RL that aims to optimize traffic efficiency, driving safety, and driving comfort simultaneously by designing a reward function that incorporates these three performance criteria. Compared to SAINT-ACC, however, Zhu does not account for dynamically changing traffic conditions effectively as the traffic efficiency is determined according to a static probability density function, and the driving safety depends on a fixed TTC$^*$. The RL framework for SAINT-ACC and Zhu are implemented using Python based on Keras and Tensorflow~\cite{abadi2016tensorflow}; It is interfaced with SUMO via Traffic Control Interface (TraCI)~\cite{wegener2008traci}. Simulations are performed with a PC running MacOS (Majove 10.14.5) with a 1.4GHz Intel Core i5 CPU and 8GB of RAM.

We consider two most common highway types, \emph{i.e.,} a highway segment with an on-ramp and that with an off-ramp. We create a 1.5km highway segment with a on/off ramp with the length of 360m. There is a 180m-long acceleration lane used by vehicles to accelerate before they reach the on-ramp. The traffic is generated and injected into the main road at a rate of 1800 veh/h/lane. A vehicle used for this simulation has the body length of 4$\sim$5m and changes lanes based on a widely adopted lane-changing model~\cite{erdmann2015sumo}. SAINT-ACC allows vehicles to take an action of controlling the inter-vehicle gap every one second in response to dynamically changing traffic conditions. For non-ACC vehicles, the default
car-following model (Krauss) with driver imperfection
parameters is used. In particular, the minimum headway distance for each vehicle is intentionally set to 0m to permit collisions in simulation for the purpose of evaluating driving safety. Also, both the maximum acceleration and deceleration of a vehicle are set to sufficiently high to evaluate driving comfort considering the fact that the maximum acceleration and deceleration to ensure driving comfort should be within the range between 2 m/s$^2$ and 3 m/s$^2$~\cite{hoberock1976survey}. The performance of SAINT-ACC is compared with Zhu and the baseline method (referred to as `Base' hereafter). In the baseline method, vehicles are not equipped with any intelligent ACC technology. We measure driving safety, traffic efficiency, and driving comfort by varying the technology penetration rate, vehicle density on the on-ramp as well as off-ramp. Here the penetration means the percentage of vehicles that are equipped with an intelligent ACC system, and the vehicle density refers to the number of vehicles on a particular road segment. Specifically, driving safety is measured based on the number of near-collisions as defined in Section~\ref{sec:motivation}; Traffic efficiency is measured using the average speed of vehicles in the highway segment; And driving comfort is measured based on the acceleration and deceleration of vehicles.

\subsection{Highway with an On-Ramp}
\label{subsection_comparison}

\begin{figure}[h]
\centering
\includegraphics[width=.99\columnwidth]{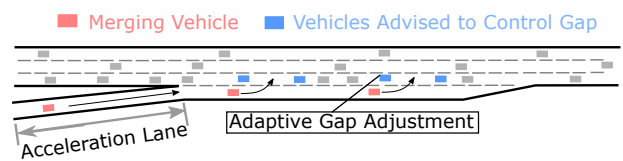}
\caption {An illustration of a highway segment with an on-ramp.}
\label{fig:onramp_scenario}
\end{figure}

We evaluate the performance on a highway segment with an on-ramp. A top view of the highway segment is depicted in Fig.~\ref{fig:onramp_scenario} that illustrates blue-colored vehicles adjust the inter-vehicle gap to allow red-colored merging vehicles to change lanes. In this simulation study, we measure the traffic efficiency, driving safety, and driving comfort of SAINT-ACC, Zhu, and Base by varying the penetration rate. We run 200 random simulations to represent a single data point.

\begin{figure}[h]
\centering
\begin{minipage}{.49\columnwidth}
  \centering
  \includegraphics[width=\textwidth]{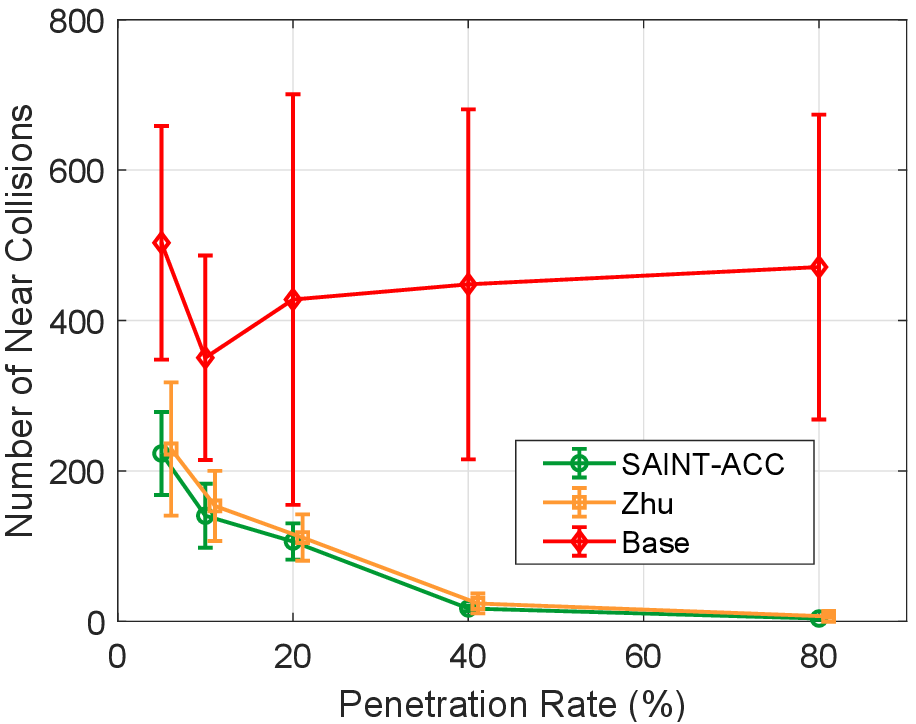}
  \vspace{-15pt}
  \caption {Driving safety for highway with on-ramp.}
  \label{fig:penetration_safety}
\end{minipage}%
\hspace*{3mm}
\begin{minipage}{.49\columnwidth}
  \centering
  \includegraphics[width=\textwidth]{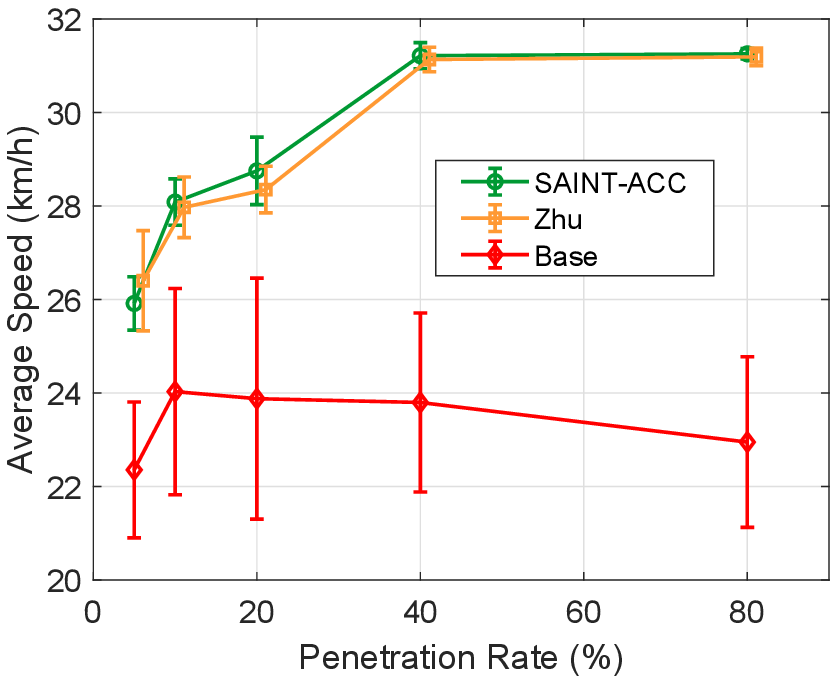}
  \vspace{-15pt}
  \caption {Traffic efficiency for highway with on-ramp.}
  \label{fig:penetration_efficiency}
\end{minipage}
\end{figure}

\begin{table}[h]
\label{tab:table_per}
\caption{Performance gain (driving safety) compared with Zhu for varying penetration rates (PR).}
\begin{center}
    \begin{tabular}{| c | c | c | c | c | c |}
    \hline
    PR & 5\% & 10\% & 20\% & 40\% & 80\% \\ \hline
    Gain & 2.6\% & 8.5\% & 7.1\% & 27.7\% & 42.4\% \\ \hline
    \end{tabular}
\end{center}
\end{table}

Fig.~\ref{fig:penetration_safety} shows the results on driving safety. No significant pattern is observed for the driving safety of Base as vehicles are not equipped with an intelligent ACC system. In contrast, SAINT-ACC and Zhu achieve significantly better driving safety compared with Base. Specifically, SAINT-ACC achieves by up to 99.2\% higher driving safety than that of Base when the penetration rate is 80\%. It is also observed that SAINT-ACC outperforms Zhu by up to 42.4\% at the penetration rate of 80\%. Table~2 summarizes the performance gain compared with Zhu for different penetration rates to supplement the graph (Fig.~\ref{fig:penetration_safety}) not clearly showing the performance gain due to the too large y-axis range. On average, the driving safety is enhanced by 17.5\% which translates into 7.2 less near collisions. Given a short period of time for each run, 7.2 less near collisions is notable as it can accumulate into a very large number over time. These results demonstrate the efficacy of adaptively adjusting TTC$^*$ and using it to control the inter-vehicle gap.

\begin{wrapfigure}{l}{0.7\columnwidth}
\vspace{-20pt}
  \begin{center}
    \includegraphics[width=\linewidth]{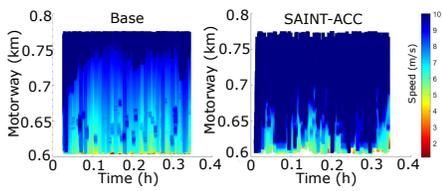}
  \vspace{-25pt}
  \caption{A space-time diagram.}
  \label{fig:timespace}
  \end{center}
  \vspace{-20pt}
\end{wrapfigure}

Fig.~\ref{fig:penetration_efficiency} depicts the results on traffic efficiency. Both SAINT-ACC and Zhu have significantly higher traffic efficiency compared with Base because they adapt the inter-vehicle gap so that merging vehicles change the lane to the mainline smoothly with reduced interruption to traffic flow of the mainline. The average speed of SAINT-ACC is higher by up to 36.2\% compared with Base when the penetration rate is 80\%. Fig.~\ref{fig:timespace} shows a time-space diagram that more clearly represents the qualitative behavior of SAINT-ACC compared with Base, suppressing formation of a traffic jam caused by merging vehicles.

\begin{figure}[h]
\centering
\begin{minipage}{.49\columnwidth}
  \centering
  \includegraphics[width=\textwidth]{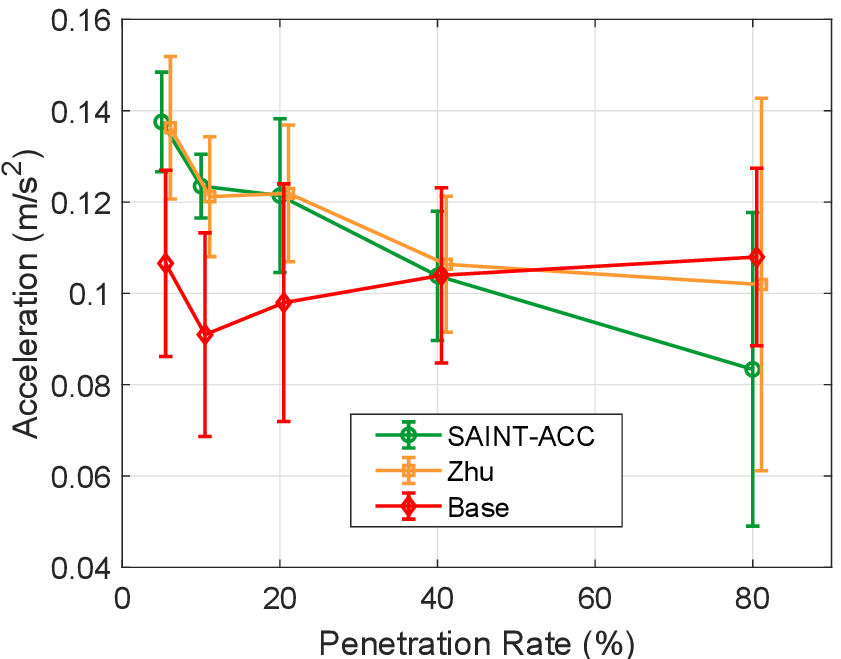}
  \vspace{-15pt}
  \caption {Driving comfort for highway with on-ramp.}
  \label{fig:penetration_comfort}
\end{minipage}%
\hspace*{3mm}
\begin{minipage}{.49\columnwidth}
\centering
\includegraphics[width=\textwidth]{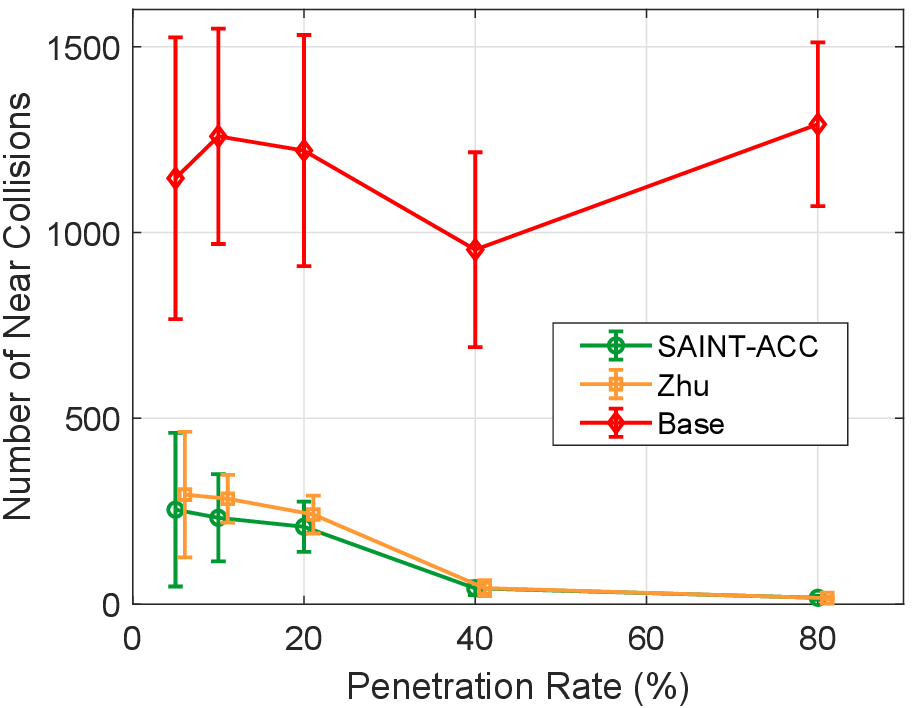}
\vspace{-15pt}
\caption {Driving safety for highway with off-ramp.}
\label{fig:penetration_safety_off}
\end{minipage}
\end{figure}


Fig.~\ref{fig:penetration_comfort} shows the results on driving comfort. The literature shows that the maximum acceleration and deceleration to assure comfortable driving is in the range between 2 m/$s^2$ and 3 m/$s^2$~\cite{hoberock1976survey}. Considering this range, we observe that driving comfort is not significantly impacted by all approaches regardless of the penetration rate. An interesting observation is that when the penetration rate is small (\emph{i.e.,} smaller than 40\%), both SAINT-ACC and Zhu exhibit lower driving comfort level compared with that of Base. The reason is possibly that when vehicles adjust the inter-vehicle gap to allow merging vehicles to change lanes, the following vehicles that are not equipped with an intelligent ACC system need to decrease their speed significantly, causing degradation of driving comfort. We also observe that driving comfort for both SAINT-ACC and Zhu keep improving with higher penetration rates. When most vehicles are equipped with an intelligent ACC system, we observe that both approaches provide much more comfortable driving experience. In particular, SAINT-ACC has 18.2\% higher driving comfort level than that of Zhu when the penetration rate is 80\%.

\begin{wrapfigure}{r}{0.5\columnwidth}
\vspace{-15pt}
  \begin{center}
    \includegraphics[width=\linewidth]{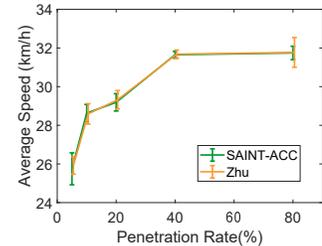}
  \vspace{-25pt}
  \caption{The behavior of the ACC agent without the TTC agent.}
  \label{fig:driving-efficiency-without-TCC}
  \end{center}
  \vspace{-12pt}
\end{wrapfigure}

Another interesting experiment is to investigate the behavior of SAINT-ACC without the TTC agent. We run the same set of simulations with and without the TTC agent. The results regarding traffic efficiency are shown in Fig.~\ref{fig:driving-efficiency-without-TCC}, demonstrating that without the dynamic adjustment of the TTC threshold, the ACC agent does not improve the performance much. Similar results are observed for driving safety and comfort.

\subsection{Highway with Off-ramp}
\label{subsec:highway_type}

\begin{figure}[h]
\centering
\includegraphics[width=.99\columnwidth]{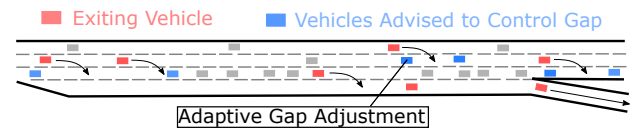}
\caption {An illustration of a highway segment with off-ramp.}
\label{fig:offramp_scenario}
\end{figure}

The performance of SAINT-ACC is measured on a highway with an off-ramp. A top view of the highway segment is depicted in Fig.~\ref{fig:offramp_scenario}. In this simulation scenario, vehicles exiting the highway are randomly selected from the three lanes of the main road. Fig.~\ref{fig:penetration_safety_off} shows the results on driving safety. Similar to the results obtained from the highway segment with an on-ramp, no significant pattern of driving safety is observed for Base since vehicles are not equipped with an intelligent ACC system. In contrast, as the penetration rate increases, driving safety is significantly increased for both SAINT-ACC and Zhu. Specifically, driving safety for SAINT-ACC is significantly higher than that of Base by up to 98.7\%. It is observed that dynamic adjustment of the inter-vehicle gap allows exiting vehicles to change lanes smoothly to exit the highway. In particular, in comparison with Zhu, we observe that SAINT-ACC improves driving safety by up to 13.8\% when the penetration rate is 5\%.

\begin{figure}[h]
\centering
\begin{minipage}{.49\columnwidth}
\centering
\includegraphics[width=\textwidth]{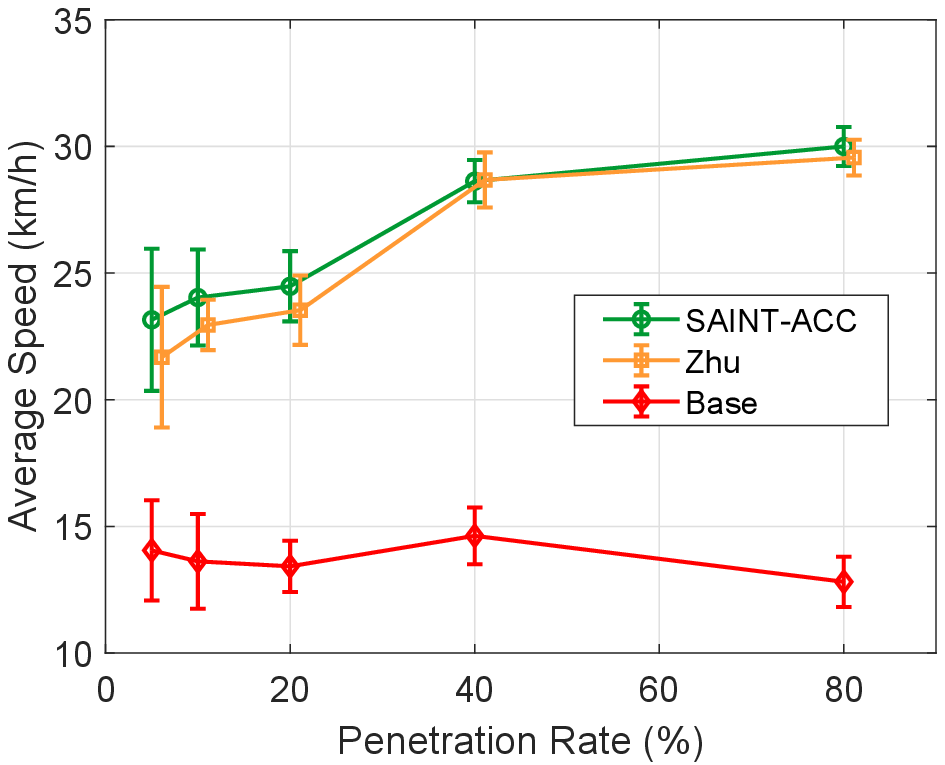}
\vspace{-15pt}
\caption {Traffic efficiency for highway with off-ramp.}
\label{fig:penetration_efficiency_off}
\end{minipage}%
\hspace*{3mm}
\begin{minipage}{.49\columnwidth}
\centering
\includegraphics[width=\textwidth]{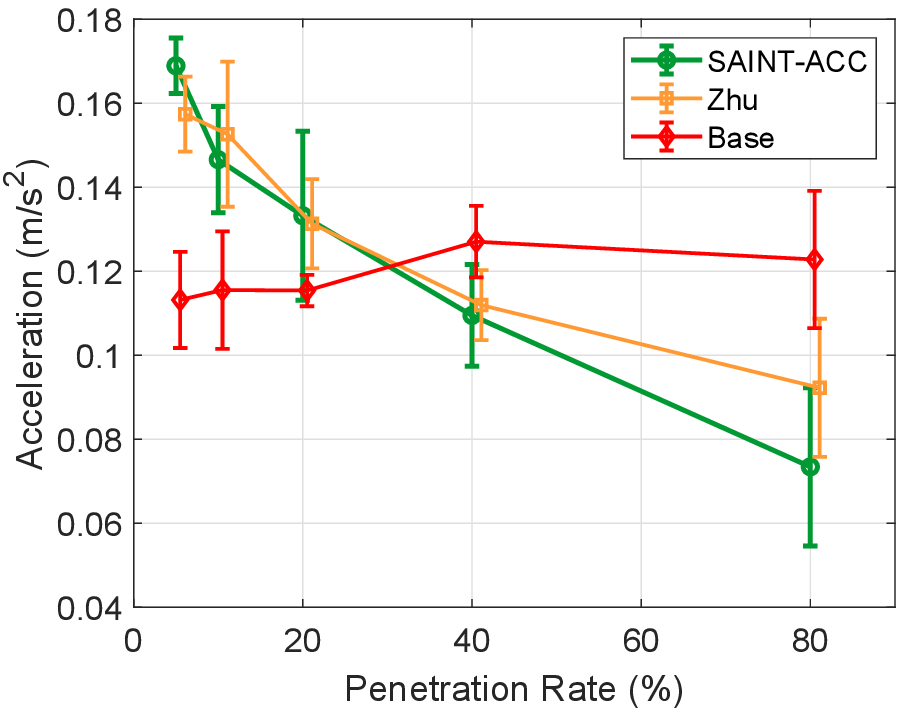}
\vspace{-15pt}
\caption {Driving comfort for highway with off-ramp.}
\label{fig:penetration_comfort_off}
\end{minipage}
\end{figure}

Fig.~\ref{fig:penetration_efficiency_off} shows the results on traffic efficiency. Both SAINT-ACC and Zhu have significantly higher traffic efficiency than Base, especially as the penetration rate increases. In particular, SAINT-ACC has higher traffic efficiency by up to 134.2\% compared with Base. This result demonstrates that SAINT-ACC enables exiting vehicles to change lanes to exit the highway without causing much interruption to surrounding vehicles since the inter-vehicle gap is appropriately adjusted. We also observe that as the penetration rate increases, the traffic efficiency of both SAINT-ACC and Zhu increases. Interestingly, the performance gap between the two approaches becomes greater when the penetration is small; It is observed that SAINT-ACC achieves higher traffic efficiency by up to 6.82\% compared with Zhu. Although the performance gap for traffic efficiency seems small, the results demonstrate the capability of SAINT-ACC achieving higher traffic efficiency while keeping superior driving safety than that of Zhu.

Fig.~\ref{fig:penetration_comfort_off} shows the results on driving comfort. Similar results are observed as those obtained from the highway with an on-ramp. Overall, the impact of exiting vehicles on driving comfort is insignificant in our simulation setting. Another interesting observation is that driving comfort for both SAINT-ACC and Zhu keeps increasing as the penetration rate increases. Compared with Base, SAINT-ACC improves the level of driving comfort by up to 40.2\% when the penetration is as high as 80\%. However, when the penetration rate is small (\emph{e.g.,} $<$ 30\%), driving comfort for SAINT-ACC becomes even worse than that for Base. We think that the reason is because the following vehicles without an intelligent ACC system have to accelerate/decelerate as the front vehicle equipped with SAINT-ACC or Zhu adjusts the inter-vehicle gap.


\subsection{Effect of Merging Traffic}
\label{sec:effect_of_traffic}

\begin{figure}[h]
\centering
\begin{minipage}{.49\columnwidth}
\centering
\includegraphics[width=\textwidth]{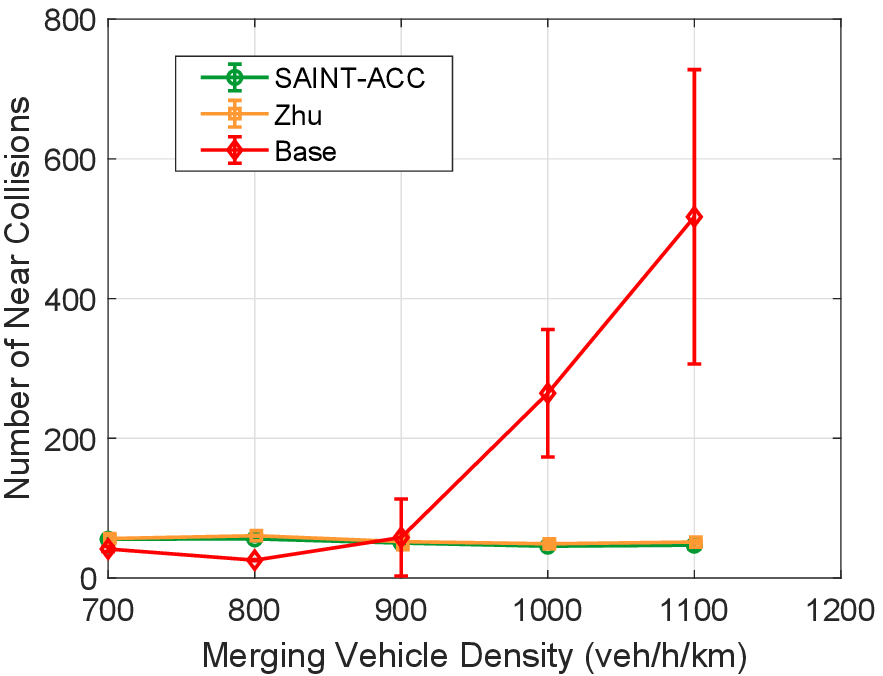}
\vspace{-15pt}
\caption {Effect of merging traffic on driving safety.}
\label{fig:merging-vehicle-density-near-collision}
\end{minipage}%
\hspace*{3mm}
\begin{minipage}{.49\columnwidth}
\centering
\includegraphics[width=\textwidth]{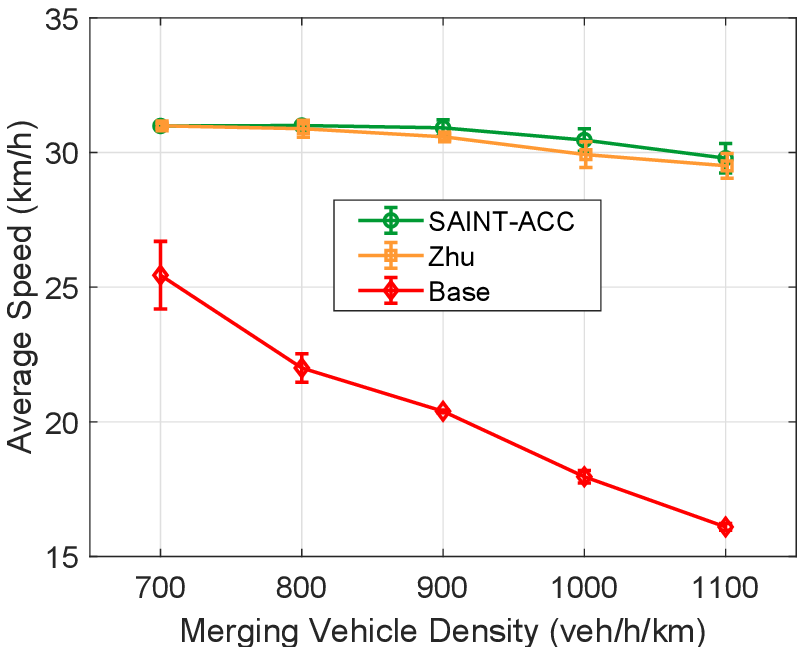}
\vspace{-15pt}
\caption {Effect of merging traffic on traffic efficiency.}
\label{fig:merging-vehicle-density-avg-speed}
\end{minipage}
\end{figure}

Merging traffic is one of the main causes of traffic congestion~\cite{milanes2010automated}. We evaluate the performance of SAINT-ACC by varying the merging traffic density defined as the number of vehicles on a merging lane per unit time (veh/h/km). Fig.~\ref{fig:merging-vehicle-density-near-collision} displays the driving safety of all three approaches as a function of the merging vehicle density. The results show that the driving safety of Base degrades when there are more merging vehicles. It is also observed that the traffic becomes increasingly unstable as the merging vehicle density increases as indicated by the large standard deviation. In contrast, both SAINT-ACC and Zhu successfully suppress near-collisions even when the merging vehicle density is very high. Specifically, SAINT-ACC improves driving safety by up to 93.6\% compared with that of Base, indicating that SAINT-ACC manages effectively complex traffic conditions with very high merging vehicle density. In comparison with Zhu, SAINT-ACC achieves higher driving safety by up to 8.7\% demonstrating the effectiveness of dynamically adjusting TTC$^*$.

Fig.~\ref{fig:merging-vehicle-density-avg-speed} shows the results on traffic efficiency. It is observed that when no intelligent ACC solution is applied, traffic efficiency degrades as the merging vehicle density increases. In contrast, both SAINT-ACC and Zhu effectively sustain high average vehicle speed despite the increasing merging vehicle density. In particular, SAINT-ACC achieves higher average speed by up to 83.9\% compared with that of Base, demonstrating the effectiveness of dynamic adaptation of the inter-vehicle gap in response to varying merging vehicle density. We also observe that SAINT-ACC achieves higher traffic efficiency compared with Zhu.

\begin{figure}[h]
\centering
\begin{minipage}{.49\columnwidth}
\centering
\includegraphics[width=\textwidth]{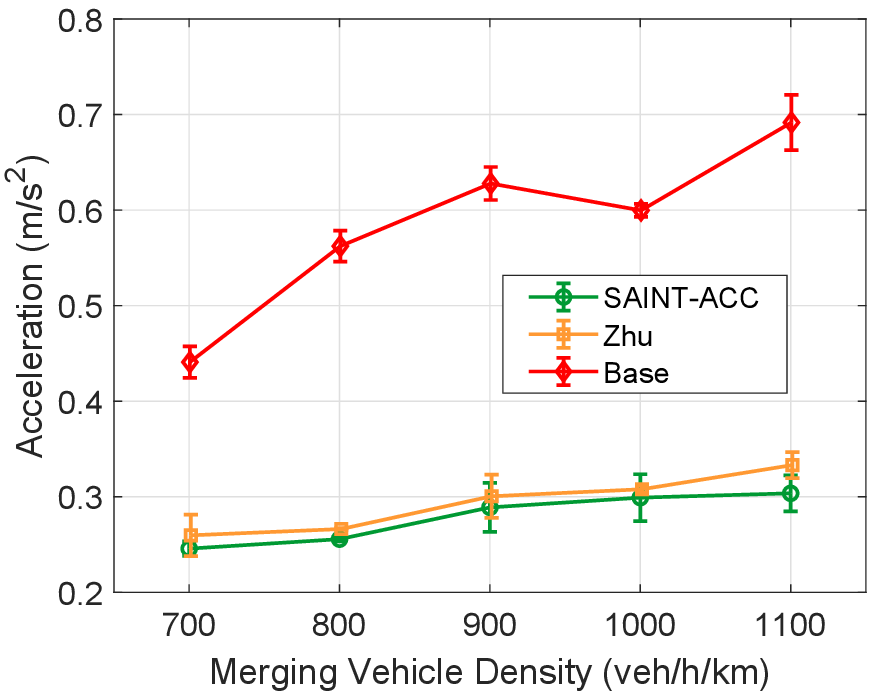}
\vspace{-15pt}
\caption {Effect of merging traffic on driving comfort.}
\label{fig:merging-vehicle-density-acceleration}
\end{minipage}%
\hspace*{3mm}
\begin{minipage}{.49\columnwidth}
\centering
\includegraphics[width=\textwidth]{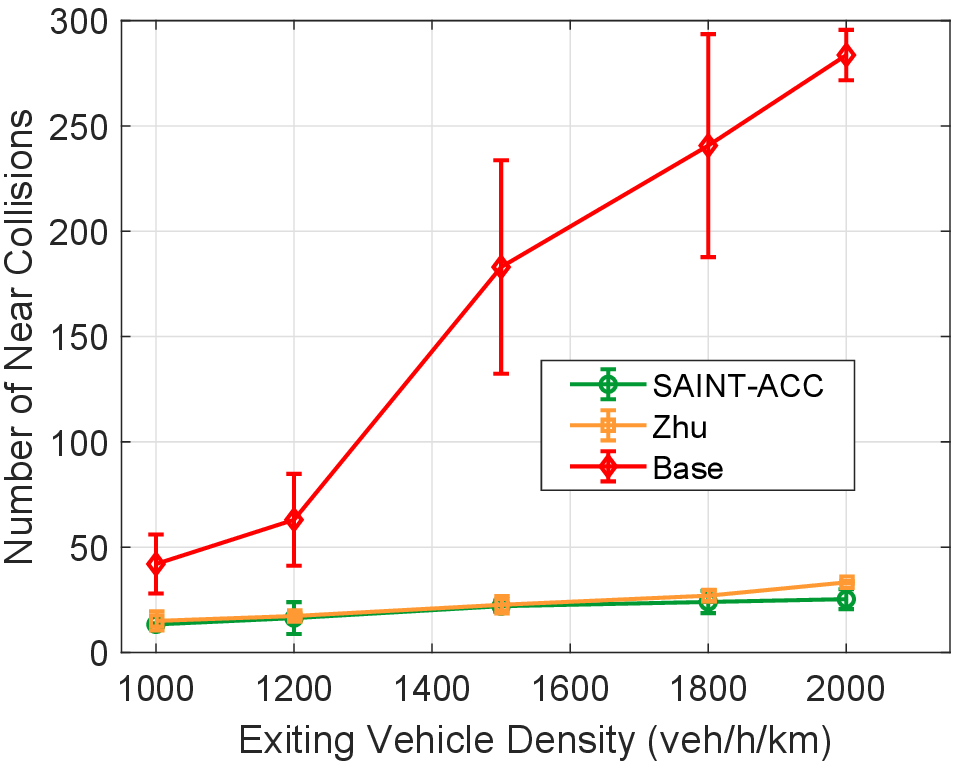}
\vspace{-15pt}
\caption {Effect of exiting traffic on driving safety.}
\label{fig:exiting-vehicle-density-near-collision}
\end{minipage}
\end{figure}

It is observed that the level of driving comfort decreases for all three approaches as the merging vehicle density increases as depicted in Fig.~\ref{fig:merging-vehicle-density-acceleration}. It is also observed that both SAINT-ACC and Zhu improve the driving comfort level compared with that of Base. However, it should be mentioned that the results are insignificant as the average acceleration of the vehicles for all approaches is within the range of the acceleration for comfortable driving~\cite{hoberock1976survey} regardless of the merging vehicle density.

\subsection{Effect of Exiting Traffic}
\label{sec:effect_of_traffic}

\begin{figure}[h]
\centering
\begin{minipage}{.49\columnwidth}
\centering
\includegraphics[width=\textwidth]{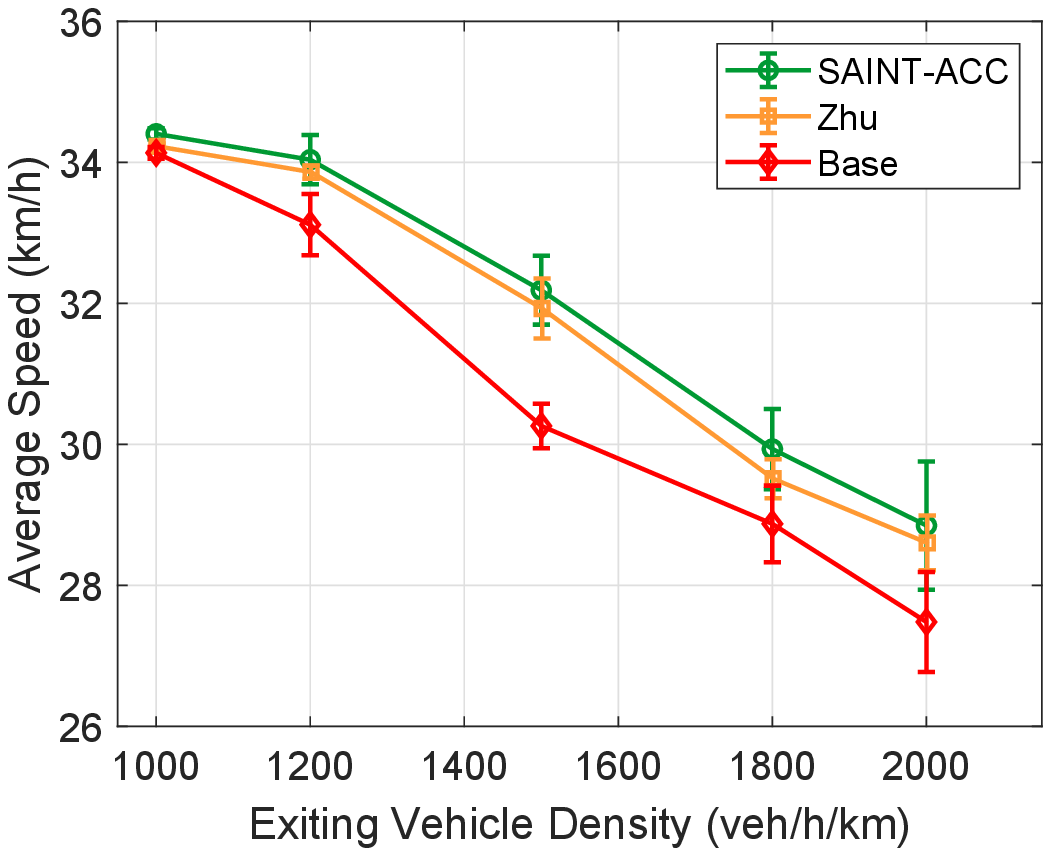}
\vspace{-15pt}
\caption {Effect of exiting traffic on traffic efficiency.}
\label{fig:exiting-vehicle-density-avg-speed}
\end{minipage}%
\hspace*{3mm}
\begin{minipage}{.49\columnwidth}
\centering
\includegraphics[width=\textwidth]{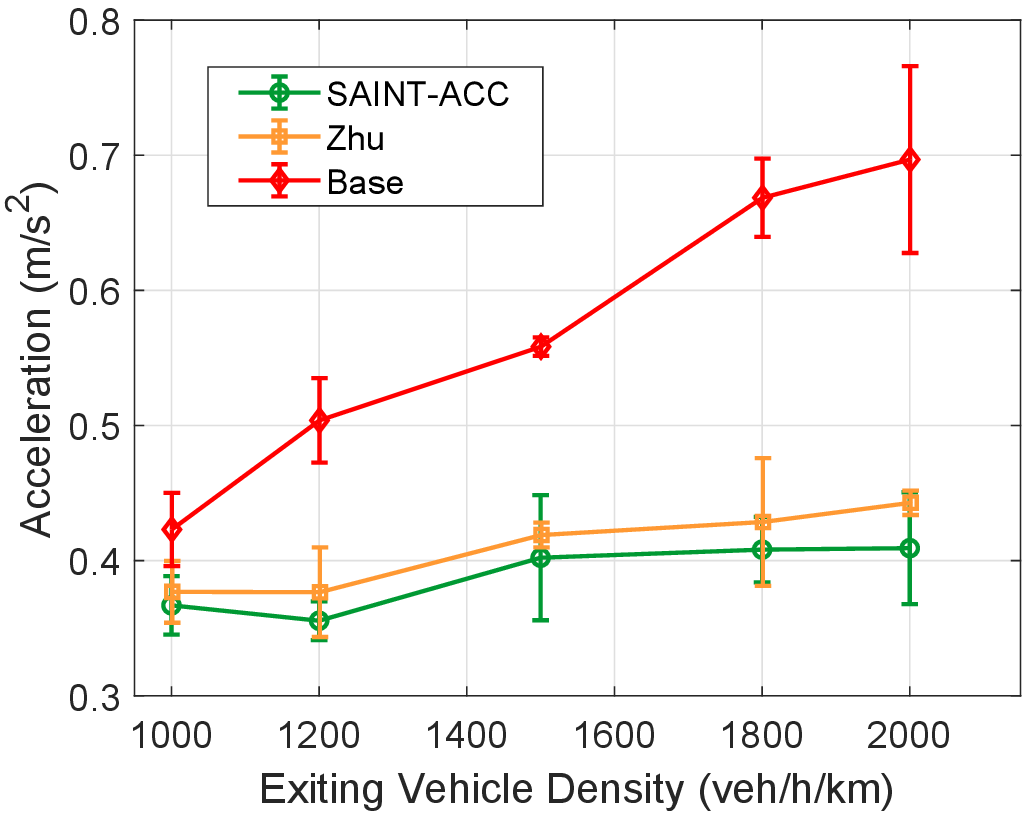}
\vspace{-15pt}
\caption {Effect of exiting traffic on driving comfort.}
\label{fig:exiting-vehicle-density-acceleration}
\end{minipage}
\end{figure}

Exiting traffic (vehicles exiting the highway) is another important cause for traffic congestion~\cite{gunther2012mitigating}. We evaluate the performance of SAINT-ACC by varying the exiting traffic density defined as the number of vehicles exiting the highway per unit time (veh/h/km). In particular, exiting vehicles are randomly selected from all three lanes of the main road in our simulation. Fig.~\ref{fig:exiting-vehicle-density-near-collision} shows the effect of exiting traffic density on driving safety. It can be seen that the driving safety of Base significantly degrades as there are more vehicles exiting the highway. In contrast, both SAINT-ACC and Zhu keep the number of near-collisions fairly small despite the increasing number of exiting vehicles. Especially, SAINT-ACC enhances driving safety by up to 91.5\% compared with that of Base. Furthermore, SAINT-ACC achieves better driving safety in comparison with that of Zhu as well by up to 24\% when the exiting vehicle density is 2,000 veh/h/km. The results clearly demonstrate the advantages of adjusting the inter-vehicle gap adaptively using the optimal TTC$^*$ in response to dynamically changing traffic conditions.

\begin{wrapfigure}{r}{0.5\columnwidth}
\vspace{-15pt}
  \begin{center}
    \includegraphics[width=\linewidth]{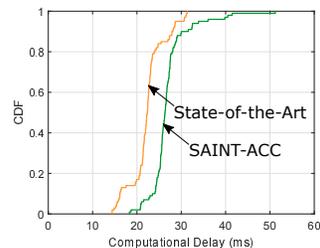}
    \vspace{-15pt}
    \caption{Computation delay.  \label{fig:computational_time}}
  \end{center}
  \vspace{-10pt}
\end{wrapfigure}

Fig.~\ref{fig:exiting-vehicle-density-avg-speed} depicts the results on traffic efficiency with varying exiting traffic density. We observe that the traffic efficiency of all three approaches decreases as the exiting vehicle density increases. Interestingly, it is observed that traffic efficiency is less significantly affected by exiting traffic compared to how merging traffic influences on traffic efficiency. A possible explanation is that vehicles exiting the highway are randomly distributed on all three lanes; as such, the traffic flow of all three lanes are affected by the exiting vehicles; In contrast, all merging vehicles attempt to change lanes from a single lane (\emph{i.e.,} a ramp), thereby affecting only the traffic flow on the lane adjacent to the ramp, creating greater traffic perturbation. Overall, compared with Base and Zhu, SAINT-ACC improves the traffic flow by up to 11.8\% and 4.1\%, respectively. Fig.~\ref{fig:exiting-vehicle-density-acceleration} displays the results on driving comfort which demonstrate that SAINT-ACC improves the level of driving comfort by up to 41.3\% and 7.6\% in comparison with that of Base and Zhu, respectively.

\subsection{Computational Delay}
\label{sec:computation_time}

We measure the computational delay required to make a control decision on the inter-vehicle gap. The computational delay is measured 100 times per epoch for SAINT-ACC and Zhu. Fig.~\ref{fig:computational_time} depicts the cumulative distribution function (CDF) graph of the computational delay. The average computational delay for Zhu and SAINT-ACC is 22.3ms and 26.9ms, respectively. It can be seen that the computational delay for SAINT-ACC is 17\% higher on average than that of Zhu. The reason is SAINT-ACC has a more complex RL model trained with a larger number of traffic parameters; additionally, SAINT-ACC is designed with dual RL models. However, the results demonstrate that the average computation delay for SAINT-ACC is still sufficiently small for vehicles to make a control decision every one-second update interval.

\subsection{Update Interval}
\label{sec:update_interval}

\begin{wrapfigure}{r}{0.5\columnwidth}
\vspace{-35pt}
  \begin{center}
    \includegraphics[width=\linewidth]{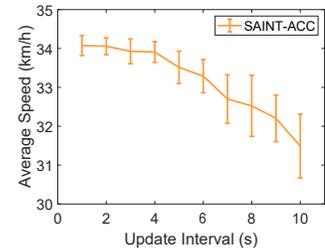}
    \vspace{-15pt}
    \caption{Effect of the update Interval of the inter-vehicle distance.  \label{fig:effect_of_interval}}
  \end{center}
  \vspace{-10pt}
\end{wrapfigure}

SAINT-ACC makes a control decision of the inter-vehicle distance every second, allowing a vehicle to cope with dynamically changing traffic conditions more effectively. Additionally, SAINT-ACC allows that the update interval can be easily changed depending on traffic situations. We evaluate the effect of the update interval on the performance of SAINT-ACC focusing on traffic efficiency. Results are depicted in Fig.~\ref{fig:effect_of_interval}. As shown, as the update interval increases, the traffic efficiency gradually degrades. The reason is straightforward that vehicles by updating the inter-vehicle distance more frequently cope better with the dynamically changing traffic conditions. Another interesting observation is that when the update interval is sufficiently small (\emph{i.e.,} $<$ 3 seconds), the performance gain is marginal for a smaller update interval, allowing us to choose an udpate interval between 1 second and 3 seconds.

\section{Conclusion}
\label{sec:conclusion}

We have presented a safety-aware intelligent ACC system that maximizes traffic efficiency, driving safety, and driving comfort simultaneously by finding and adjusting the optimal TTC threshold used to control the inter-vehicle gap adaptively in response to dynamically changing traffic conditions. Our future work is to implement a proof-of-concept SAINT-ACC system on a real vehicle platform and perform experiments to validate the performance under real-world traffic environments using open-source semi-autonomous driving systems such as Comma.AI~\cite{santana2016learning}.



\bibliography{output}
\bibliographystyle{icml2021}

\end{document}